\documentclass[10pt,twocolumn,letterpaper]{article}

\usepackage{iccv}
\usepackage{times}
\usepackage{epsfig}
\usepackage{graphicx}
\usepackage{amsmath,amssymb,amsfonts}
\usepackage{hyphenat}
\usepackage[table]{xcolor}
\usepackage[font=small,belowskip=0pt, aboveskip=6pt]{caption}
\usepackage{multirow}
\usepackage{booktabs,amsfonts,dcolumn}
\usepackage{tabularx}
\usepackage{makecell}
\usepackage{hyperref}
\usepackage{comment}
\usepackage{diagbox}

\usepackage[numbers]{natbib}
\usepackage{multicol}
\usepackage{amsmath,amssymb,amsfonts}
\usepackage[ruled,linesnumbered]{algorithm2e}
\usepackage{graphicx}
\usepackage{multirow}
\usepackage{soul}
\usepackage{siunitx}
\usepackage{physunits}
\usepackage{xspace}

\usepackage{physics}
\usepackage{amsmath}
\usepackage{tikz}
\usepackage{mathdots}
\usepackage{yhmath}
\usepackage{cancel}
\usepackage{color}
\usepackage{siunitx}
\usepackage{array}
\usepackage{multirow}
\usepackage{gensymb}
\usepackage{tabularx}
\usepackage{extarrows}
\usepackage{booktabs}
\usetikzlibrary{fadings}
\usetikzlibrary{patterns}
\usetikzlibrary{shadows.blur}
\usetikzlibrary{shapes}

\usepackage{subcaption}
\usepackage{enumitem}
\usepackage{gensymb}
\makeatletter
\DeclareRobustCommand\onedot{\futurelet\@let@token\@onedot}
\def\@onedot{\ifx\@let@token.\else.\null\fi\xspace}

\def\eg{\emph{e.g}\onedot}

\def\etal{\emph{et al}\onedot}
\makeatother

\iccvfinalcopy 

\def\iccvPaperID{****} 

\begin{document}

\title{\LARGE \bf
Word2Wave: Language Driven Mission Programming for\\ Efficient Subsea Deployments of Marine Robots%
\vspace{-5mm}
}
\author{Ruo Chen, David Blow, Adnan Abdullah, and Md Jahidul Islam\\
{\small Email: {\tt \{chenruo@,david.blow@,adnanabdullah@,jahid@ece.\}ufl.edu}} \\
{
\small RoboPI Laboratory, Department of ECE, University of Florida, FL 32611, USA. 
} 
\thanks{This pre-print is accepted for publication at ICRA-2025.} 
\thanks{Info: ~\url{https://robopi.ece.ufl.edu/word2wave.html}}
\vspace{-3mm}
}

\maketitle

\textit{\textbf{Abstract.} This paper explores the design and development of a language-based interface for dynamic mission programming of autonomous underwater vehicles (AUVs). The proposed `Word2Wave' (W2W) framework enables interactive programming and parameter configuration of AUVs for remote subsea missions. The W2W framework includes: (i) a set of novel language rules and command structures for efficient language-to-mission mapping; (ii) a GPT-based prompt engineering module for training data generation; (iii) a small language model (SLM)-based sequence-to-sequence learning pipeline for mission command generation from human speech or text; and (iv) a novel user interface for 2D mission map visualization and human-machine interfacing. The proposed learning pipeline adapts an SLM named T5-Small that can learn language-to-mission mapping from processed language data effectively, providing robust and efficient performance. In addition to a benchmark evaluation with state-of-the-art, we conduct a user interaction study to demonstrate the effectiveness of W2W over commercial AUV programming interfaces. Across participants, W2W-based programming required less than 10\% time for mission programming compared to traditional interfaces; it is deemed to be a simpler and more natural paradigm for subsea mission programming with a usability score of 76.25. W2W opens up promising future research opportunities on hands-free AUV mission programming for efficient subsea deployments.}

\section{Introduction}
Recent advancements in Large Language Models (LLMs) and speech-based human-machine dialogue frameworks are poised to revolutionize robotics by enabling more natural and spontaneous interactions~\cite{wang2024large,zhang2023large}. These systems allow robots to interpret natural human language for more accessible and interactive operation, especially in challenging field robotics applications. In particular, remote deployments of autonomous robots in subsea inspection, surveillance, and search and rescue operations require dynamic mission adjustments on the fly~\cite{islam2018dynamic}. Seamless mission parameter adaptation and fast AUV (Autonomous Underwater Vehicle) deployment routines are essential features for these applications, which the traditional interfaces often fail to ensure~\cite{lucas2020survey,sattar2009underwater,islam2018understanding, antonio2017Nav}.

\begin{figure}[t]
     \centering
     \begin{subfigure}[]{0.45\textwidth}
         \centering
         \includegraphics[width=\linewidth]{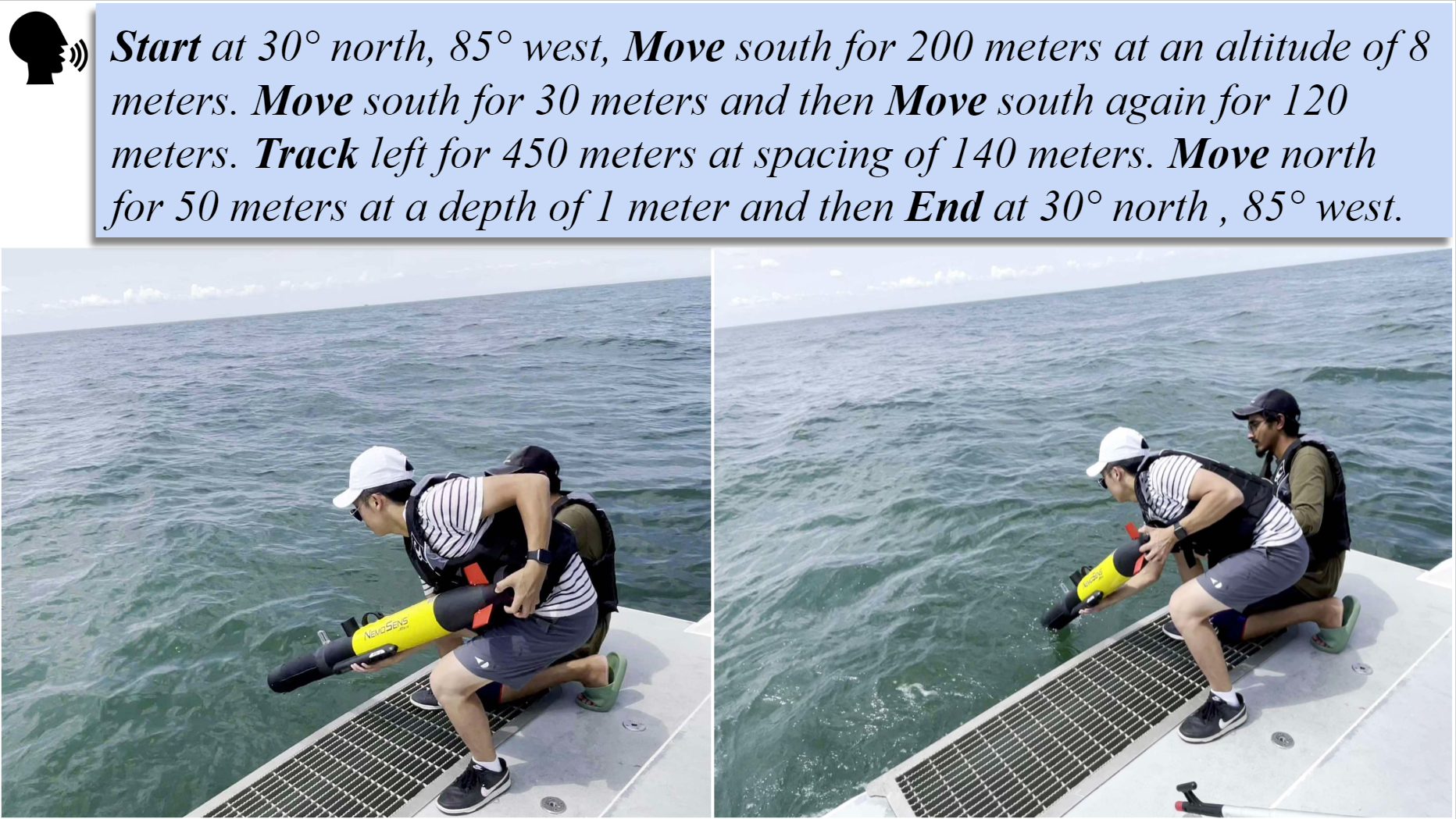}%
         \vspace{-1mm}
         \caption{A W2W-programmed mission and subsea deployment scenario.}
         \label{fig1a}
     \end{subfigure}
     \vspace{1mm}
     
     \begin{subfigure}[]{0.45\textwidth}
         \centering
         \includegraphics[width=\linewidth]{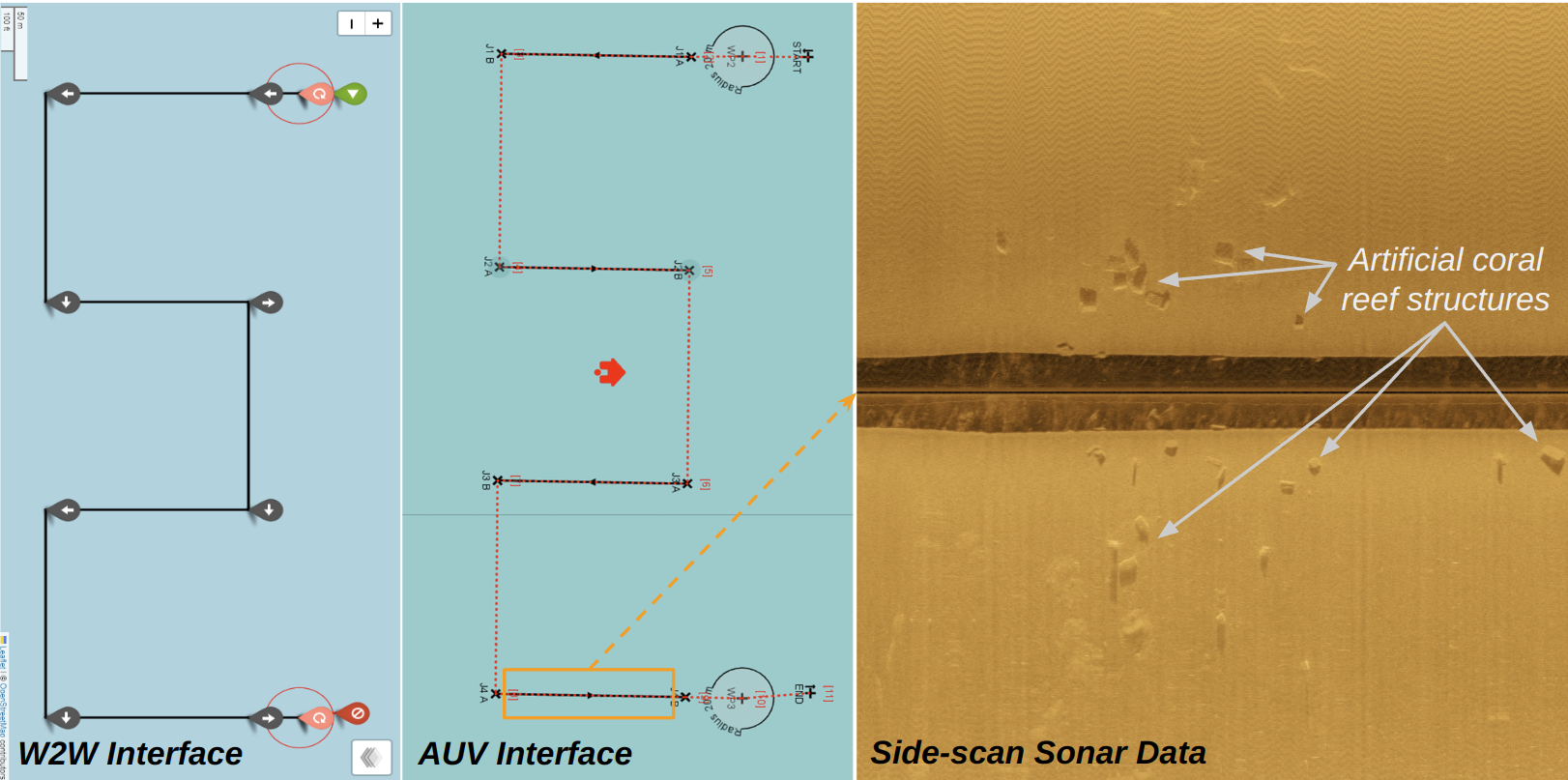}%
         \vspace{-1mm}
         \caption{Corresponding mission maps and sample sensory data.}
         \label{fig1b}
     \end{subfigure}
     \vspace{-1mm}
        \caption{The proposed \textbf{Word2Wave (W2W)} framework offers a novel user interface for real-time AUV mission programming with natural human language. We demonstrate the effectiveness of W2W over traditional interfaces by subsea mission deployments on a NemoSens AUV platform.}%
        \label{fig:beauty_shot}
\vspace{-4mm}
\end{figure}%

The existing subsea robotics technologies offer predefined planners that require manual configuration of mission parameters in a complex software interface~\cite{gussen2021optimization,zhou2023review}. It is extremely challenging and tedious to program complex missions spontaneously, even for skilled technicians, especially on an undulating vessel when time is of the essence. Natural language-based interfaces have the potential to address these limitations by making mission programming more user-friendly and efficient. Recent works have shown how the reasoning capabilities of LLMs can be applied for mission planning~\cite{Kong2024_EmbodiedAI} and human-robot interaction (HRI)~\cite{wang2024large,lin2023development} by using deep vision-language
models~\cite{park2023visual,shao2023prompting}, text-to-action paradigms~\cite{brooks2023instructpix2pix,huang2022inner}, and embodied reasoning pipelines~\cite{lin2023development,wang2024large}. Contemporary research demonstrates promising results~\cite{yang2023oceanchat,bi2023oceangpt,lucas2020survey} on language-based human-machine interfaces (HMIs) for subsea mission programming as well.  

In this paper, we introduce ``Word2Wave", a small language model (SLM)-based framework for real-time AUV programming in subsea missions. It provides an interactive HMI that uses natural human speech patterns to generate subsea mission plans to perform autonomous remote operations. For implementation, we adapt an SLM training pipeline based on the Text-To-Text Transfer Transformer (T5)~\cite{Raffel2019ExploringTL} small model ({\tt T5-Small}) to parse natural human speech into a sequence of computer-interpretable commands. These commands are subsequently converted to a set of waypoints and mission milestones for an AUV to execute. 


We design the language rules of Word2Wave (W2W) with seven atomic commands to support a wide range of subsea mission plans, particularly focusing on subsea surveying, mapping, and inspection tasks~\cite{andy2022subsea}. These commands are simple and intuitive, yet powerful tools, to program complex missions without using tedious software interfaces of commercial AUVs~\cite{NemoSens,mcmahon2016mission}. W2W includes all basic operations to program widely used mission patterns such as \textit{lawnmower}, \textit{spiral}, \textit{ripple}, and \textit{polygonal} trajectories at various configurations. With these flexible operations, W2W allows users to program subsea missions using natural language, similar to how they would describe the mission to a human diver. To this end, we designed a GPT~\cite{achiam2023gpt}-based prompt engineering module~\cite{palnitkar2023chatsim} for comprehensive training data generation. 

The proposed learning pipeline demonstrates a delicate balance between robustness and efficiency, making it ideal for real-time use. Through comprehensive quantitative and qualitative assessments, we demonstrate that SLMs can capture targeted vocabulary from limited data with an adapted learning pipeline and articulated language rules. Specifically, our adapted {\tt T5-Small} model provides state-of-the-art (SOTA) performance for sequence-to-sequence learning while offering $2\times$ faster inference rate (of $70.5$\,ms) with $85.1\%$ fewer parameters than computationally demanding LLMs such as BART-Large~\cite{bartlarge}. We also investigated other SLM architectures (\eg, MarianMT~\cite{mariannmt}) for benchmark evaluation based on accuracy and computational efficiency.


Moreover, we develop an interactive user interface (UI) for translating the W2W-generated language tokens into 2D \textit{mission maps}. Unlike traditional HMIs, it adopts a minimalist design intended for futuristic use cases. Specifically, we envision that users will engage in interactive dialogues for formulating and planning subsea missions. While such HMIs are still an open problem, our proposed UI is significantly more efficient and user-friendly -- which we validate by a thorough user interaction study with $15$ participants.

From the user study, we find that on average, participants took less than $10\%$ time for mission programming by W2W compared to using traditional interfaces. The participants, especially those with prior experiences of subsea deployments, preferred using W2W as a simpler and more intuitive programming paradigm. They rated W2W with a usability score~\cite{brooke1996sus} of $76.25$, validating that it induces less cognitive load and requires minimal technical support for novice users. With these features, the proposed W2W framework takes us a step forward to our overarching goal of integrating human-machine dialogue for embodied reasoning and hands-free mission programming of marine robots.

\section{Background And Related Work}
\subsection{Language Models for Human-Machine Embodied Reasoning and HRI}
\vspace{-1mm}
Classical language-based systems focus on deterministic human commands for controlling mobile robots~\cite{chandarana2017fly,garcia2019high}. Traditionally, the open-world navigation with visual goals (ViNGs)~\cite{shah2021ving} or visual-inertial navigation (VIN)~\cite{huang2019visual} pipelines have been mostly independent of human language inputs~\cite{chandarana2018challenges,shah2011language,mutschler2005language}. In these systems, language/speech is parsed separately as a control input to the ViNG or VIN systems to achieve planning~\cite{shah2011language,silva2014development} and navigation~\cite{trujillo2017using,cividanes2021extended}.

With the advent of LLMs, contemporary robotics research has focused on leveraging the power of natural language for human-robot embodied decision-making and shared autonomy~\cite{lin2023development,wang2024large,Kong2024_EmbodiedAI,Cui2024_UAVTaskPlanning}. A key advancement is the development of vision-language
models (VLMs)~\cite{park2023visual,shao2023prompting} for human-machine embodied reasoning. Huang~\etal~\cite{huang2022inner} developed an ``Inner Monologue" framework that injects sensorimotor feedback
into a LLM, which prompts as the robot interacts with the environment. In ``InstructPix2Pix"~\cite{brooks2023instructpix2pix}, Tim~\etal~combines an LLM and a text-to-image model for visual question answering. 
These features allow robots to understand visual content with relevant linguistic information, as shown by Wu~\etal~in the ``TidyBot" system~\cite{wu2023tidybot}.

While general-purpose LLMs are resource intensive, Small Language Models (SLMs) are often more suited for targeted robotics applications~\cite{Nazarov2025_SLM,kwon2024language}, especially for when computational resources are constrained~\cite{Thawakar2024MobiLlama}. SLMs in various zero-shot learning pipelines have been fine-tuned for applications such as long-horizon navigation~\cite{shah2023lm}, embodied manipulation~\cite{gao2024physically}, and trajectory planning~\cite{kwon2024language,mandi2024roco}. These are emerging technologies and, thus, ongoing developments for more challenging real-world field robotics applications. 


\subsection{Subsea Mission Programming Interfaces}
\vspace{-1mm}
Leveraging human expertise is critical for configuring subsea mission parameters because fully autonomous mission planning and navigation are challenging underwater~\cite{sanchez2020autonomous,wei2021architecture,mcmanus2005multidisciplinary,oliveira1998mission}. Human-programmed missions enable AUVs to adapt to dynamic mission objectives and deal with environmental challenges with no GPS or wireless connectivity~\cite{gonzalez2020autonomous,whitt2020future}. Various HRI frameworks~\cite{sattar2009underwater,islam2018understanding,enan2022robotic}, telerobotics consoles~\cite{abdullah2024ego2exo,xu2021vr}, and language-based interfaces~\cite{hallin2009using,mcmahon2016mission} have been developed for mission programming and parameter reconfiguration. For instance, visual languages such as ``RoboChat"~\cite{Dudek2007_RoboChat} and ``RoboChatGest"~\cite{Xu2008_RoboChatGest} use a sequence of symbolic patterns to communicate simple instructions via {AR-Tag markers} and hand gestures, respectively. These and other language paradigms~\cite{park2023visual} are mainly suited for short-term human-robot cooperative missions. 

For subsea telerobotics, augmented and virtual reality (AR/VR) interfaces integrated on traditional consoles are gaining popularity in recent times~\cite{Capocci2017_ROVReview,VideoRay}. These offer immersive teleop~\cite{xu2024augmented,laranjeira20203d} and improve teleoperators' perception of environmental semantics~\cite{islam2024eob,gao2020mission,manley2018aquanaut,parekh2023underwater,samuelson2024guided}. Long-term autonomous missions are generally planned offline in terms of a sequence of waypoints following a specific trajectory~\cite{gao2020mission,lucas2020survey,rankin2021robotic} such as lawn mower, perimeter following, fixed-altitude spiral/corkscrew patterns, polygons, etc. Yang~\etal~\cite{yang2023oceanchat} proposed an LLM-driven OceanChat system for AUV motion planning in HoloOcean simulator~\cite{potokar2022holoocean}. Despite inspiring results in marine simulators~\cite{palnitkar2023chatsim,manhaes2016uuv,prats2012open}, the power of language models for subsea mission deployments~\cite{bi2023oceangpt} on real systems is not explored in depth.

\section{Word2Wave: Language Design}

\subsection{Language Commands And Rules}\label{lang_rules}
\vspace{-1mm}
Designing mission programming languages for subsea robots involves unique considerations due to the particular requirements of marine robotics applications. In Word2Wave, our primary objective is to enable spontaneous human-machine interaction by natural language. We also want to ensure that the high-level abstractions in the Word2Wave (W2W) language integrate with the existing industrial interfaces and simulation environments for seamless adaptations. 

\begin{table}[h]
\centering
\caption{The language commands, parameters, and command structure notations of Word2Wave are shown.}
\label{tab:TokenTable}
\vspace{-1mm}
\small
\scalebox{0.80}{
\begin{tabular}{l||l|l}
\Xhline{2\arrayrulewidth}
\cellcolor{gray!10}\textbf{Parameters} & \multicolumn{2}{l}{{\tt b}: bearing; ~{\tt d}: depth; ~{\tt s}: speed} \\
 & \multicolumn{2}{l}{{\tt a}: altitude; ~{\tt t}: turns; ~{\tt r}:  
    radius; {\tt n}: no change}   \\
 & \multicolumn{2}{l}{{\tt tab}: spacing; ~{\tt dir}: direction; ~{\tt dist}: distance} \\
 & \multicolumn{2}{l}{{\tt cw, ccw}: clockwise, counter clockwise} \\
\Xhline{2\arrayrulewidth}
    \cellcolor{gray!10}\textbf{Command} & \cellcolor{gray!10}\textbf{Symbol} & \cellcolor{gray!10}\textbf{Language Structure} \\ \Xhline{2\arrayrulewidth}
    {Start/End} & {\tt S/E} & \tt [S/E: latitude, longitude] \\  \hline
    {Move} & {\tt Mv} & \tt [Mv: b, d, s, d/a/n, d/a(m)] \\ \hline
    {Track} & {\tt Tr} & \tt [Tr: dir, tab, end, d/a/n, d/a(m)] \\ \hline
    {Adjust} & {\tt Az} & \tt [Az: d/a, d/a(m)] \\ \hline
    {Circle} & {\tt Cr} & \tt [Cr: t, r, cw/ccw, d/a/n, d/a(m)] \\ \hline
    {Spiral} & {\tt SP} & \tt [SP: t, r, cw/ccw, d/a/n, d/a(m)] \\
\Xhline{2\arrayrulewidth}
\end{tabular}
}
\vspace{-2mm}
\end{table}

As shown in Table~\ref{tab:TokenTable}, we consider $7$ language commands in {Word2Wave}. 
Their intended use cases are as follows.
\begin{enumerate}[label={$\arabic*$.},nolistsep,leftmargin=*]
\item \textbf{Start}/\textbf{End} ({\tt S/E}): is intended to {start} or {end} a mission at a given \textit{latitude} and \textit{longitude}. These coordinates are two input parameters, taken in decimal degrees with respect to true North and West, respectively.

\item \textbf{Move ({\tt Mv})}: command is designed to move the AUV to a specified \textit{distance} (meters) at a given \textit{bearing} (w.r.t. North) and \textit{speed} (m/s). If/when the speed is not explicitly stated, a default value of $1$\,m/s is used.

\item \textbf{Track ({\tt Tr})}: is used to plan a set of parallel lines given a \textit{direction} orthogonal to the current bearing (same altitude), \textit{spacing} between each line, and ending \textit{distance} in meters. 

\item \textbf{Adjust ({\tt Az})}: generates a waypoint to adjust the AUV's target depth or altitude at its current location.
        
\item \textbf{Circle ({\tt Cr})}: creates a waypoint commanding the AUV to circle around a position for several \textit{turns} at a given \textit{radius} in a clockwise or counterclockwise \textit{direction}.

\item \textbf{Spiral ({\tt Sp})}: generates a circular pattern that starts at a central point and then expands outwards over a series of \textit{turns} out to a specific \textit{radius}. The spiral \textit{direction} is set to either clockwise or counterclockwise.
\end{enumerate}
\vspace{1mm}

\begin{figure}[t]
     \centering
     \begin{subfigure}[]{0.5\textwidth}
         \centering         \includegraphics[width=\linewidth]{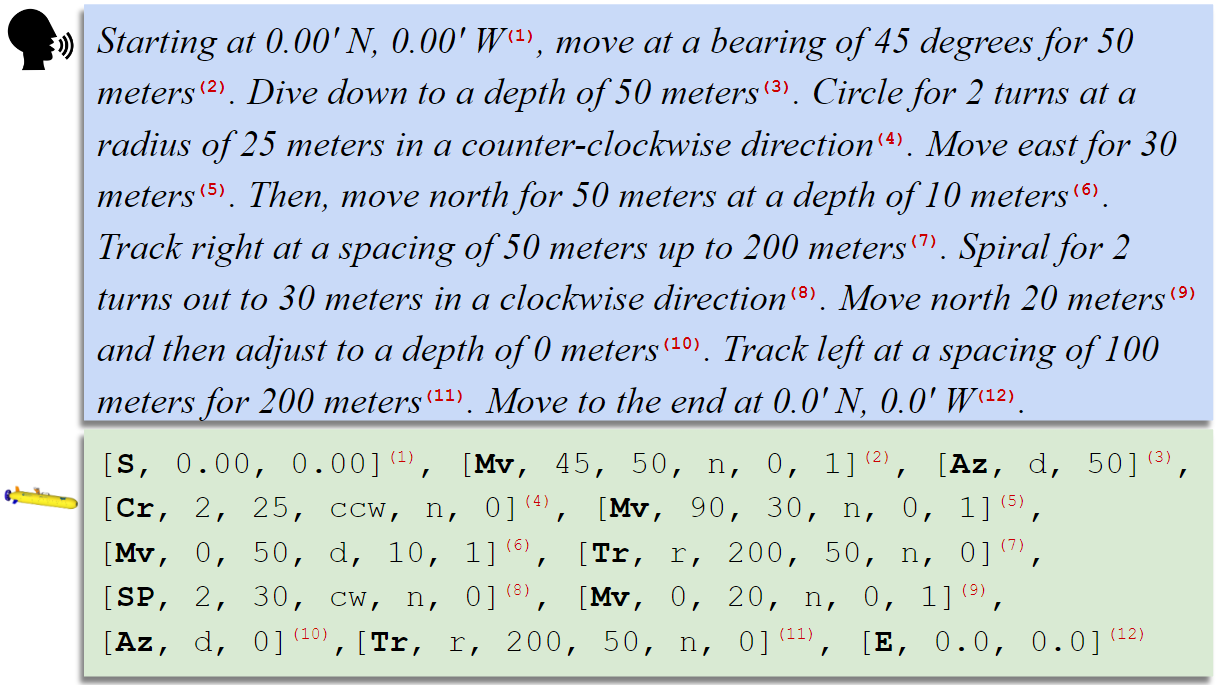}%
         \vspace{-1.5mm}
         \caption{An example of speech being translated into a series of language tokens pertaining to the mission waypoints.}
         \label{fig:SpeechtoMission}
     \end{subfigure}
     \vspace{1mm}
     
     \begin{subfigure}[]{0.5\textwidth}
         \centering
         \includegraphics[width=\linewidth]{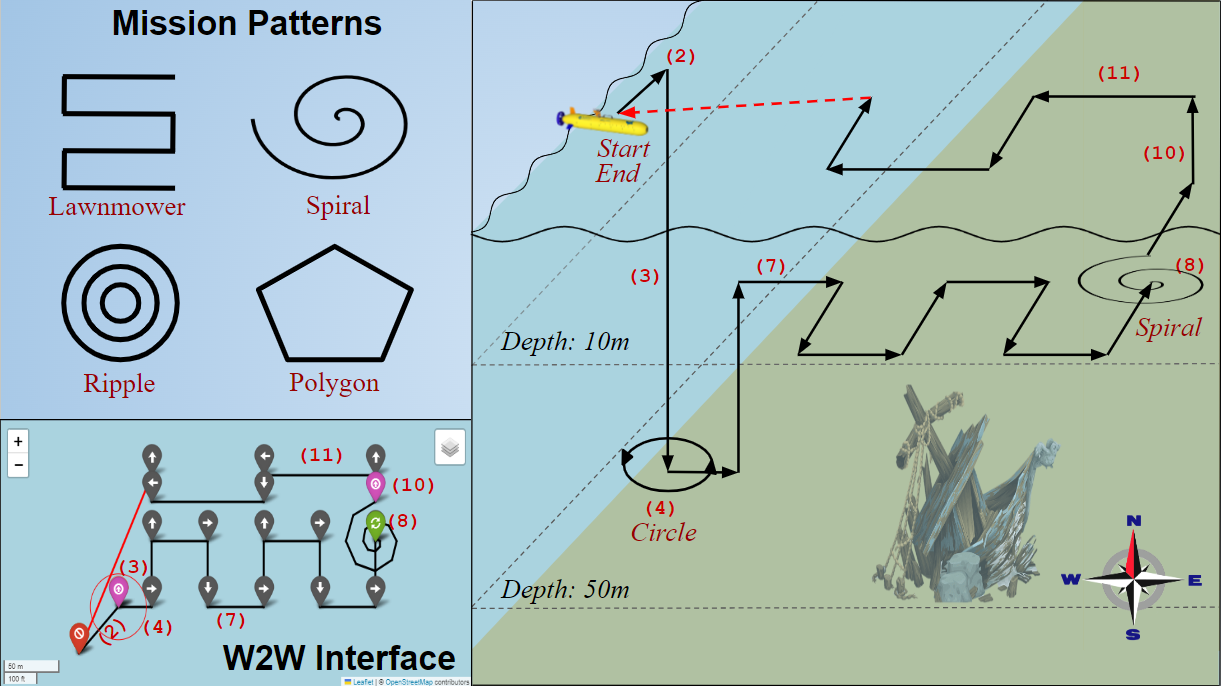}%
         \vspace{-1.5mm}
         \caption{A mission sketch and corresponding W2W-generated map are visualized for the waypoint sequence shown in (a).}
         \label{fig:MissionExample}
     \end{subfigure}
     \vspace{-1mm}
        \caption{Demonstration of a sample mission programmed using the proposed Word2Wave (W2W) interface.}
        \label{fig:system_example}
        \vspace{-5mm}
\end{figure}

If a change in depth or altitude is omitted in the input, then there is considered \textit{no change} ({\tt n}). This results in the most recent setting of depth/altitude being retained; see Fig.~\ref{fig:system_example}.

\vspace{0mm}
\subsection{Mission Types And Parameter Selection}
\vspace{-1mm}
W2W can generate arbitrary mission patterns with varying complexity using the language commands of Table~\ref{tab:TokenTable}. The only limitations are the maneuver capabilities of the host AUV. We demonstrate a particular mission programming instance in Fig.~\ref{fig:SpeechtoMission}. Mission parameters for each movement command are chosen based on commonly used terms to ensure compatibility; we particularly explore four most widely used mission patterns, as shown in Fig.~\ref{fig:MissionExample}. These atomic patterns can be further combined to plan multi-phase composite missions for a given application scenario. 
\begin{enumerate}[label={$\bullet$},nolistsep,leftmargin=*]
\item \textbf{Lawnmower} (also known as boustrophedon) patterns are ideal for surveying large areas over subsea structures. It is not suited for missions requiring more intricate movements or targeted actions. These patterns are most commonly used for sonar-based mapping, as they offer even coverage with a simple and efficient route over a large area.
\item \textbf{Polygonal} routes are best suited for irregular terrain as they provide more precise control over the AUV trajectory. They are more flexible as polygons can be tailored towards specific mission requirements and adapt to local terrains. Hence, they are generally used for missions involving inspecting known landmarks or targeted waypoints. 
\item \textbf{Ripple} patterns are defined as a series of concentric circles with either or both varying radii and depths. These are better suited when coverage is needed over a specific area and when equal spacing is important when collecting data. They are best suited for missions where sampling in varying depths of the water column is required.
\item \textbf{Spiral} paths allow for more concentrated coverage over a specific area. It operates similarly to the ripple pattern but allows for a smooth, continuous trajectory that either radiates outwards or converges to a specific point. Spiral paths do not contain well-defined boundaries compared to ripple patterns but do offer some energy savings due to having continuous motion with smaller changes to trajectory. These are commonly used for search missions requiring high-resolution coverage of a specific point.
\end{enumerate}

\vspace{0mm}

\subsection{UI For Language To Mission Mapping}
\vspace{-0mm}
Real-time visual feedback is essential for interactive verification of language to mission translation. To achieve this, we develop a 2D mission visualizer in W2W that places the human language commands into mission paths on a map for confirmation. Specifically, we generate a \textit{Leaflet map}~\cite{Leaflet}-based UI on the given GPS coordinates; we further integrate options for subsequent interactions on the map, such as icons, zoom level, and map movement or corrections.

\begin{figure}[h]
\centering
\includegraphics[width=\linewidth]{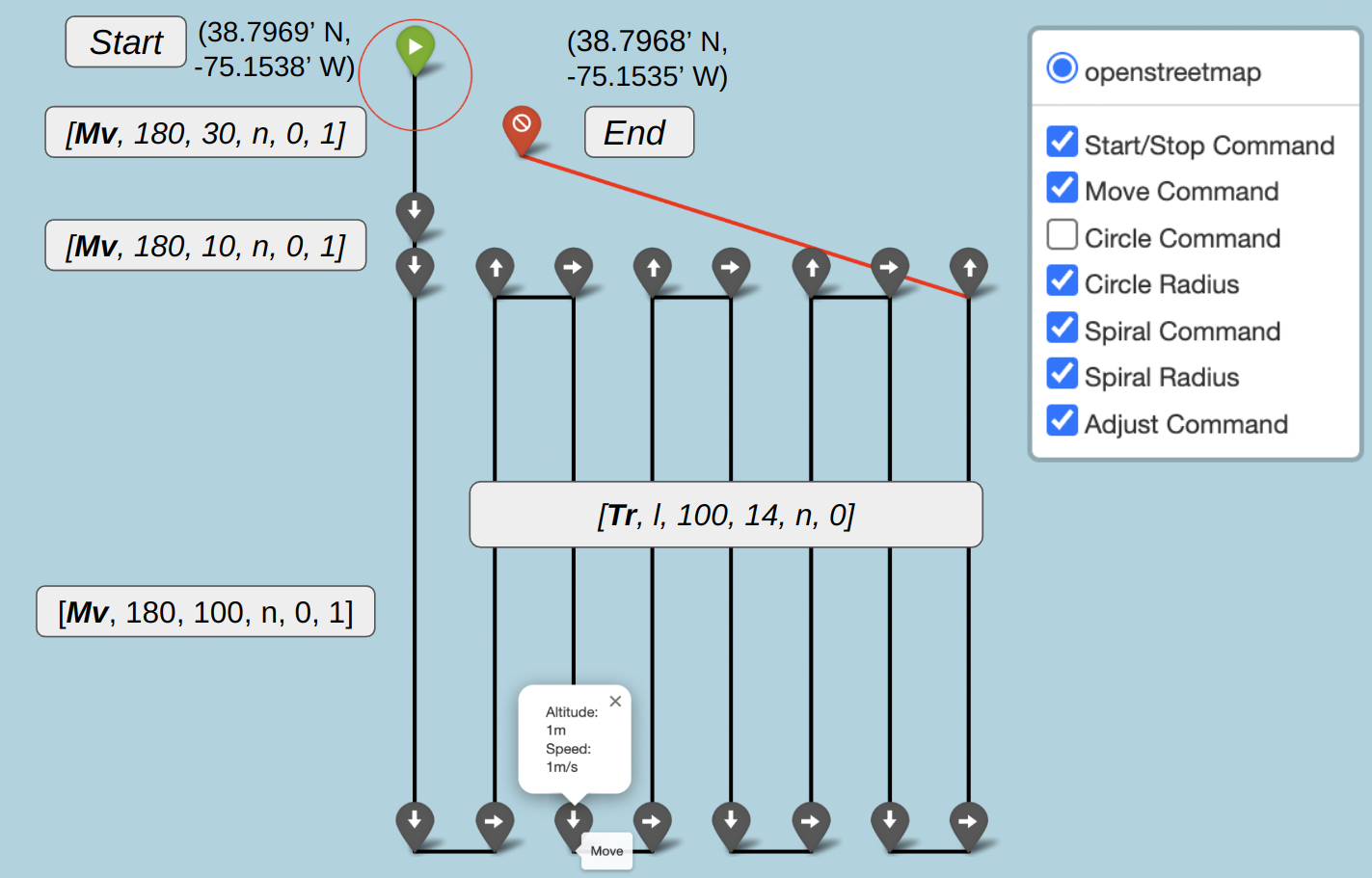}

\includegraphics[width=\linewidth]{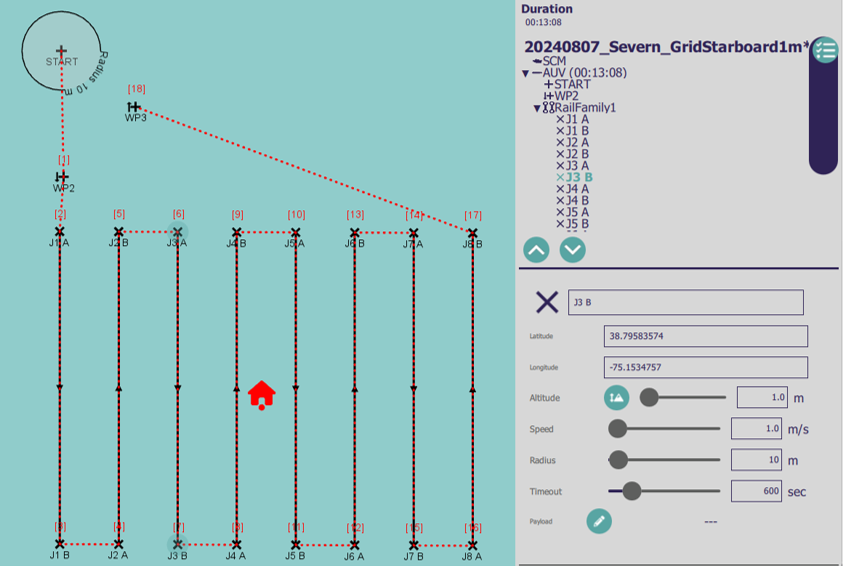}%
\vspace{0mm}
\caption{The W2W-generated mission map (top) and the corresponding map loaded on the AUV mission interface (bottom) are visualized; The W2W mission text used to generate this is: \textcolor{blue!70}{``\textit{\textbf{Start} at $38.7969\degree$\,N, $75.1538\degree$\,W, \textbf{Circle} for a turn at a radius of $10$\,m in a clockwise direction at an altitude of $1$\,m. \textbf{Move} south $30$\,m and then \textbf{Move} south $10$\,m. \textbf{Move} south for $100$\,m and then \textbf{Track} left for $100$\,m at a spacing of $14$\,m. \textbf{End} at $38.7968\degree$\,N, $75.1535\degree$\,W}''}.}
\vspace{0mm}
\label{fig:MissionMap}
\end{figure}

Fig.~\ref{fig:MissionMap} shows a sample W2W-generated map with icons representing the corresponding command tokens. All individual tokens can be visualized on separate layers for further mission adaptations by the user. The corresponding map on the AUV interface is also shown in Fig.~\ref{fig:MissionMap}; it offers the waypoints with additional information. Upon confirmation, this mission map can be loaded onto the AUV for deployment.

\section{Word2Wave: Model Training}
\subsection{Model Selection and Data Preparation}\label{sec:model_select}
\vspace{-1mm}
Structured mission command generation from an input paragraph is effectively a text-to-text translation task. Thus, we consider SLM pipelines that are effective for text translation. While popular architectures such as GPT and BERT are mainly suited for text generation and classification~\cite{achiam2023gpt,devlin2018bert}, \textit{Seq2Seq} (Sequence to Sequence) models are more suited for translation tasks~\cite{TransformerBasedNewsSummarization}. For this reason, we choose the T5 architecture, a \textit{Seq2Seq} model which provides SOTA performance in targeted `{text-to-text}' translation tasks~\cite{TransformerBasedNewsSummarization}.

In particular, we adopt a smaller variant of the original T5, named {\tt T5-Small} as it is lightweight and computationally efficient~\cite{Raffel2019ExploringTL,kwon2024language}. It processes the input text by encoding the sequence and subsequently decoding it into a concise and coherent summary. We leverage its pre-trained weights and fine-tune it on W2W translation tasks following the language rules and command structures presented in Sec.~\ref{lang_rules}.

\vspace{1mm}
\noindent
\textbf{Dataset Generation}. LLM/SLM-based subsea mission programming for marine robotics is a relatively new area of research. There are no large-scale datasets available for generalized model training. Instead, we take a prompt engineering approach using OpenAI's ChatGPT-4o~\cite{achiam2023gpt}, which helps produce various ways to program a particular mission. Different manners of phrasing the same command within the prompt ensure that various speech patterns are captured for comprehensive training. Particular attention is given so that outputs from the prompt follow the structure of a valid mission. A total of $1110$ different mission samples are prepared for supervised training. We verified each sample and their paired commands to ensure that they represent valid subsea missions for AUV deployment.






\subsection{Training And Hyperparameter Tuning}\label{sec:training_tuning}
\begin{figure}[h]
     \centering
     {\includegraphics[width=\columnwidth]{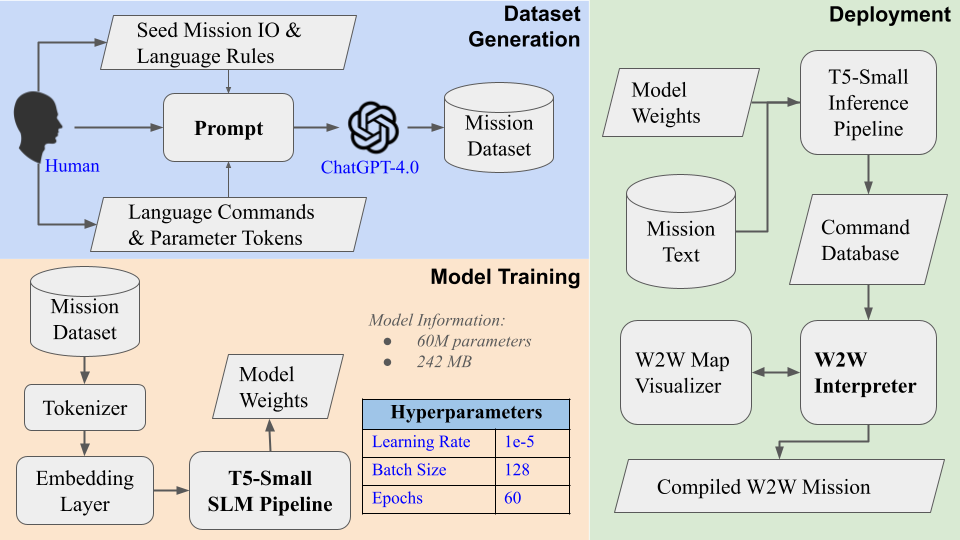}}%
     \vspace{-1mm}
     \caption{Outlines for dataset preparation, model training, and deployment processes of the W2W framework are shown.}%
     \vspace{-0mm}
     \label{fig:train_deployment}
 \end{figure}

\vspace{-1mm}
We use a randomized $80$-$5$ percentile split for training and validation; the remaining $15\%$ samples are used for testing. During training, each line in the dataset is inserted into a tokenizer to extract embedded information. The training is conducted over $60$ epochs on a machine with $32$\,GB of RAM and a single Nvidia RTX $3060$ Ti GPU with $8$\,GB of memory. The training is halted when the validation loss consistently remains below a threshold of $0.2$. The training process and hyperparameter information is shown in Fig.~\ref{fig:train_deployment}. Note that our training pipeline is not intended for general-purpose text-to-text translation but rather targeted learning on a specific set of mission vocabulary. We make sure that the supervised learning strictly adhered to the language structure of the mission commands outlined in Sec.~\ref{lang_rules}. 

\section{Experimental Analyses}
\label{sec:Experimental}

\subsection{Language Model Evaluation}
\vspace{-1mm}
Language models differ significantly in performance for certain tasks based on their underlying architectures, learning objectives, and application-specific design choices. As mentioned earlier, we adapt a sequence-to-sequence learning pipeline based on the {\tt T5-Small} model~\cite{Raffel2019ExploringTL}. For performance baseline and ablation experiments, we investigate two other SOTA models of the same genre: {BART-Large}~\cite{bartlarge} and {MarianMT}~\cite{mariannmt}. We compare their performance based on model accuracy, robustness, and computational efficiency.



\vspace{1mm}
\noindent
\textbf{Evaluation metrics and setup.} To perform this analysis in an identical setup, we train and validate each model on the same dataset until validation loss reaches a plateau. For accuracy, we use two metrics: {BLEU} (bilingual evaluation understudy)~\cite{BLEU} and {METEOR} (evaluation of translation with explicit ordering)~\cite{METEOR}. BLEU offers a language-independent understanding of how close a predicted sentence is to its ground truth. METEOR additionally considers the order of words; it evaluates the translated output based on the harmonic mean of unigram precision and recall. In addition, we compare their inference rates on $200$ test samples on the same device (RTX $3060$\,TI GPU with $32$\,GB memory). It is measured as the average time taken to generate a complete mission map given the input sample. Lastly, model sizes are apparent from the respective computational graphs.

In addition, we compare their inference rates on $200$ test samples on the same device (RTX $3060$\,TI GPU with $32$\,GB memory). It is measured as the average time taken to generate a complete mission map given the input sample. Lastly, model sizes are apparent from the respective computational graphs. 




\begin{table}[h]
\vspace{-1mm}
    \centering
    \caption{Performance comparison based on the number of parameters (in millions), BLEU and METEOR scores (in \{0,1\}), and inference speed (in miliseconds).}%
    \vspace{-1mm}
    \label{tab:model_comparison}
    \small
    \scalebox{0.75}{
    \begin{tabular}{l||r|c|c|r}
        \Xhline{2\arrayrulewidth}
        \cellcolor{gray!10}\textbf{Models} &
        \cellcolor{gray!10}\textbf{Params} ($\downarrow$)&
        \cellcolor{gray!10}\textbf{BLEU} ($\uparrow$)&
        \cellcolor{gray!10}\textbf{METEOR} ($\uparrow$)&
        \cellcolor{gray!10}\textbf{Speed}  ($\downarrow$) \\ 
        \Xhline{2\arrayrulewidth}
        {BART-Large}~\cite{bartlarge} & 406.291 M & 0.268 & 0.405 & 5122.7 ms \\ 
        {MarianMT}~\cite{mariannmt} & 73.886 M & 0.913 & 0.773 & 136.7 ms \\ 
        Ours ({T5-Small}) & 60.506 M & 0.879 & 0.813 & 70.5 ms \\ 
        \Xhline{2\arrayrulewidth}
    \end{tabular}
    }
    \vspace{-2mm}
\end{table}

\vspace{1mm}
\noindent
\textbf{Quantitative performance analyses of SOTA}. Table~\ref{tab:model_comparison} summarizes the quantitative performance comparison; it demonstrates that the MarianMT and {\tt T5-Small} models offer more accurate and consistent scores when trained on a targeted dataset compared to BART-Large. We hypothesize that LLMs like BART-Large need more comprehensive datasets and are suited for general-purpose learning. On the other hand, {\tt T5-Small} has marginally lower BLEU scores compared to MarianMT, while it offers better METEOR values at a significantly faster inference rate. {\tt T5-Small} only has about $60$\,M parameters, offering almost $2\times$ faster runtime while ensuring comparable accuracy and robustness as MarianMT. Further inspection reveals that MarianMT often randomizes the order of generated mission commands. {\tt T5-Small} does not suffer from these issues, demonstrating a better balance between robustness and efficiency. 



\begin{figure}[h]
     \centering
     {\includegraphics[width=\columnwidth]{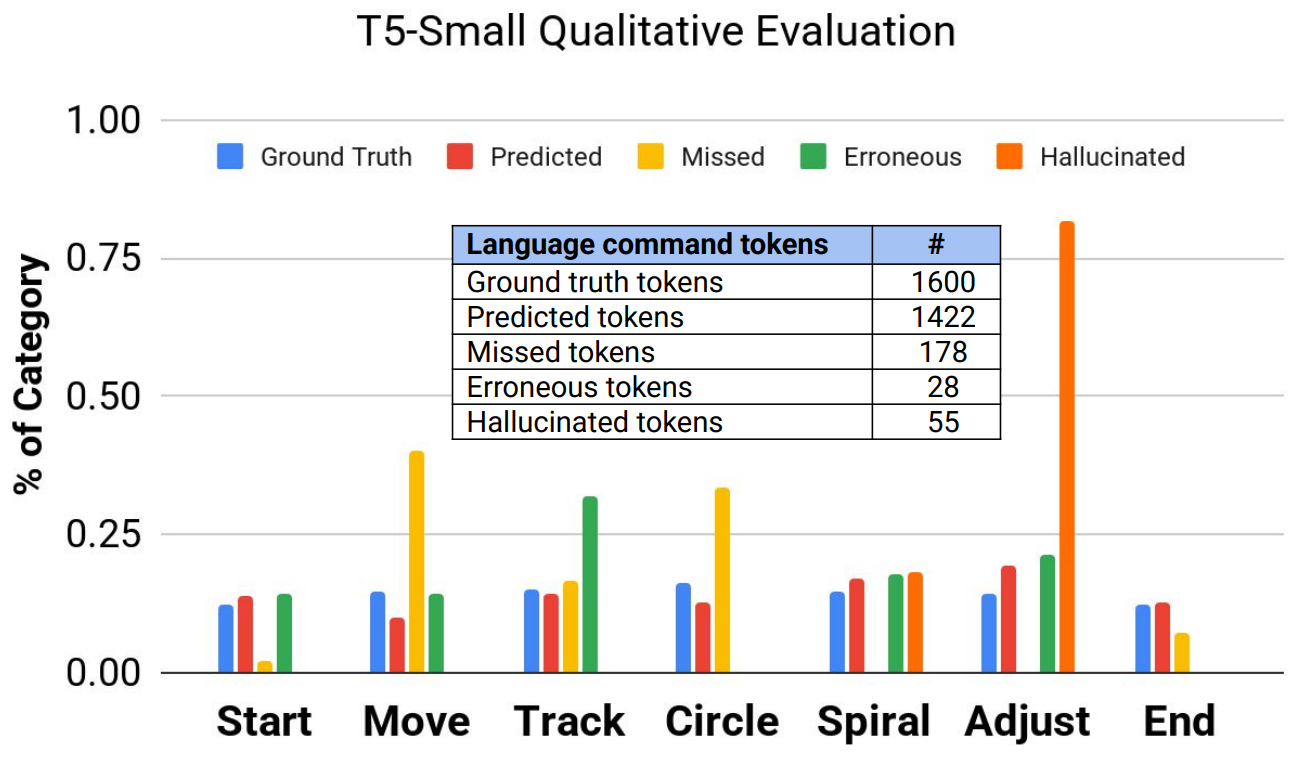}}%
     \vspace{-1mm}
     \caption{Qualitative performance of our {\tt T5-Small} mission generation engine for the $7$ language commands in W2W. 
     }%
     \vspace{-3mm}
    \label{fig:token_chart}
\end{figure}

\vspace{1mm}
\noindent
\textbf{Qualitative performance analyses}. We evaluate the qualitative outputs of {\tt T5-Small} for W2W language commands based on the number of inaccurate tokens generated across the whole test set of $200$ samples. As Fig.~\ref{fig:token_chart} shows, we categorize these into: \textit{missed} tokens (failed to generate), \textit{erroneous} tokens (incorrectly generated), and \textit{hallucinated} (extraneous) tokens. Of all the error types, we found that the \textbf{Adjust} commands are hallucinated at a disproportionately greater rate. We hypothesize that it happens due to some bias learned by the model, causing it to associate changes in depth or altitude as an \textbf{Adjust} command. This is a common issue with language models where overgeneralization outputs plausible but non-functional commands~\cite{Huang2025_LLMHallucinations}. Besides, we observed relatively high missed tokens for \textbf{Move} and \textbf{Circle} commands; \textbf{Track} is another challenging command that suffers from high error rates. Nevertheless, $89$\% tokens are accurately parsed from unseen examples, which we found to be enough for real-time mission programming.

\subsection{User Interaction Study}
\vspace{-1mm}
\noindent
We assess the usability benefits of W2W compared to the NemoSens AUV interface, which we consider as the baseline. A total of $15$ individuals between the ages of $18$ to $36$ participated in our study; $3$ of them were familiar with subsea mission programming and deployments, whereas the other $12$ people had no prior AUV programming experience. For programming with W2W, we use the VOSK speech-to-text model \cite{2024Vosk} to generate inputs, which is commonly used by mobile devices for speech transcription.

\begin{figure*}[th]
\centering
\includegraphics[width=0.4905\linewidth]{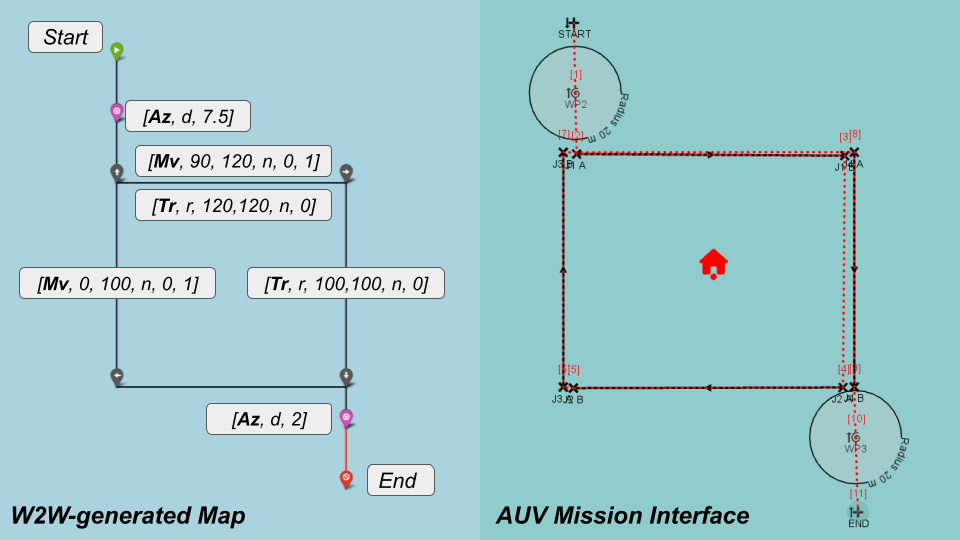}~
\includegraphics[width=0.488\linewidth]{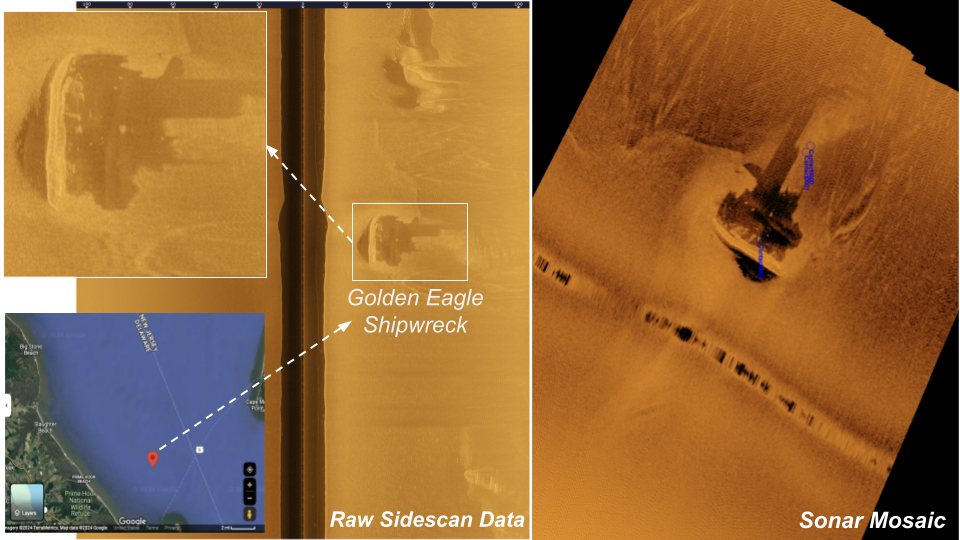}%
\vspace{-1mm}
\caption{The mission maps (left) and its sonar data (right) for a deployment at Delaware Bay (Golden Eagle shipwreck). The W2W mission command is: ``\textit{\textbf{Start} at $38.8670\degree$\,N, $75.1356\degree$\,W, \textbf{Move} south for $30$m, \textbf{Adjust} to a depth of $7.5$m. \textbf{Move} south $30$\,m and then \textbf{Move} east $120$\,m. \textbf{Track} right for $100$\,m with a spacing of $100$\,m and then \textbf{Move} north for $100$\,m. \textbf{Track} right for $120$\,m with a spacing of $120$\,m. \textbf{Move} south $20$\,m and then \textbf{Adjust} to a depth of 2m. \textbf{End} at $38.867\degree$\,N, $75.1642\degree$\,W.}''.}
\label{fig:GoldenEagle}
\vspace{-3mm}
\end{figure*}

\vspace{1mm}
\noindent
\textbf{Evaluation procedure}. Individuals were first introduced to both the W2W UI and Nemosens AUV interface for subsea mission programming. Then, they are asked to program three separate missions on these interfaces. As a quantitative measure, the total time taken to program each mission was recorded; they also completed a user survey to evaluate the degree of \textit{user satisfaction} and \textit{ease of use} between the two interfaces. This survey form was based on the system usability scale (SUS)~\cite{brooke1996sus}; the results are in Table~\ref{tab:multiprogram}.

\vspace{1mm}
\noindent
\textbf{User preference analyses}. The participants rated W2W's system usability at $76.25$, more than twice the SUS score of the baseline AUV programming interface. The participants generally expressed that the baseline interface is more complex, thus asked for assistance repeatedly. They reported several standard deviations higher scores for various features of W2W. They took less than $10\%$ time for programming missions on W2W, validating that it is more user-friendly and easy to use, compared to traditional interface.

\begin{table}[h]
    \centering
    \caption{A SUS (system usability scale)~\cite{brooke1996sus} evaluation is conducted for $10$ usability questions, scaled from 1 (strongly disagree) to 5 (strongly agree). A total of $15$ user responses are compiled as mean (std. deviation) for our W2W interface and the NemoSens AUV interface (as baseline).}%
    \small
    \vspace{-1mm}
    \scalebox{0.75}{
    \begin{tabular}{c|c|c||c|c|c}
        \Xhline{2\arrayrulewidth}
        \multicolumn{6}{l}{\textbf{Q1:} I think that I would like to use this system frequently.} \\
        \multicolumn{6}{l}{\textbf{Q2:} I found the system unnecessarily complex.} \\
        \multicolumn{6}{l}{\textbf{Q3:} I thought the system was easy to use.} \\
        \multicolumn{6}{l}{\textbf{Q4:} I would need the support of a technical person to be able to use this system.} \\
        \multicolumn{6}{l}{\textbf{Q5:} I found the various functions in this system were well integrated.} \\
        \multicolumn{6}{l}{\textbf{Q6:} I thought there was too much inconsistency in this system.} \\
        \multicolumn{6}{l}{\textbf{Q7:} Most people would learn to use this system very quickly.} \\
        \multicolumn{6}{l}{\textbf{Q8:} I found the system very cumbersome to use.} \\
        \multicolumn{6}{l}{\textbf{Q9:} I felt very confident using the system.} \\
        \multicolumn{6}{l}{\textbf{Q10:} I needed to learn a lot of things before I could get going with this system.} \\
        \Xhline{2\arrayrulewidth}
        {Q\#} &
        \cellcolor{gray!10}\textbf{Baseline} &
        \cellcolor{gray!10}\textbf{W2W Interface} &
        {Q\#} &
        \cellcolor{gray!10}\textbf{Baseline} &
        \cellcolor{gray!10}\textbf{W2W Interface} \\
        \Xhline{2\arrayrulewidth}
        1 & 2.5 (1.4) & 4.1 (0.8) & 6 & 3.7 (0.9) & 2.9 (1.0)\\
        2 & 3.4 (0.9) & 1.5 (0.8) & 7 & 2.8 (1.0) & 4.4 (1.2)\\
        3 & 2.4 (0.9) & 4.4 (0.8) & 8 & 3.1 (1.2) & 1.6 (0.9)\\
        4 & 4.0 (1.5) & 1.9 (0.7) & 9 & 2.6 (1.0) & 3.6 (1.2)\\
        5 & 2.4 (0.8) & 3.7 (1.0) & 10 & 3.5 (0.9) & 1.8 (1.0)\\ 
        \Xhline{2\arrayrulewidth}
        \multicolumn{2}{c|}{Metric} & 
        \multicolumn{2}{c|}{\cellcolor{gray!10}\textbf{Baseline}} & 
        \multicolumn{2}{c}{\cellcolor{gray!10}\textbf{W2W Interface}} \\ 
        \Xhline{2\arrayrulewidth}
        \multicolumn{2}{l|}{System usability} & 
        \multicolumn{2}{c|}{$37.5$ ($16.98$)} & 
        \multicolumn{2}{c}{$76.25$ ($19.48$)} \\ 
        \multicolumn{2}{l|}{Time taken (min:sec)} & 
        \multicolumn{2}{c|}{$27$:$02$ ($0$:$32$)} & 
        \multicolumn{2}{c}{$02$:$37$ ($0$:$03$)} \\
        \Xhline{2\arrayrulewidth}
    \end{tabular}
    }
\label{tab:multiprogram}
\vspace{-4mm}
\end{table}







\subsection{Subsea Deployment Scenarios}
\vspace{-1mm}
We use a NemoSens AUV~\cite{NemoSens} for subsea mission deployments. It is a torpedo-shaped single-thruster AUV equipped with a DVL wayfinder, a down-facing HD camera, and a $450$\,KHz side-scan sonar. As mentioned, the integrated software interface allows us to program various missions and generate waypoints for the AUV to execute in real-time. 

In this general practice, W2W interface is intended as an HMI bridge, transferring the human language into the mission map; the rest of the experimental scenarios remain the same. The mission vocabulary and language rules adopted in W2W correspond to valid subsea missions programmed on an actual robot platform. We demonstrate this with several examples from our subsea deployments in the GoM (Gulf of Mexico) and the Delaware Bay, Atlantic Ocean. Real deployments are shown in Fig.~\ref{fig:GoldenEagle} and earlier in Fig.~\ref{fig:beauty_shot}.


\vspace{5mm}
\noindent
\textbf{Limitations and future work}. Despite the accuracy and ease of use, W2W is a single-shot language summarizer, lacking human-machine \textit{dialogue} capabilities. Currently, the only way to validate mission safety is manual evaluation by the operator. We are working on extending the mission programming engine to allow multi-shot generative features to enable \textit{memory} and \textit{dialogues}. Our overarching goal is to be able to seek suggestions such as, ``\textit{We want to map a shipwreck at $38.8670\degree$\,N, $75.1356\degree$\,W location. The wreck is about $200$ ft long, the altitude clearance is $50$ ft. Suggest a few mission patterns for mapping the shipwreck}". Users may engage in subsequent dialogue to find the best mission for their intended application, taking advantage of the W2W engine's knowledge from comprehensive mission databases.

\vspace{-1mm}
\section{Conclusions}
\vspace{-1mm}
This paper presents an SLM-driven mission programming framework named Word2Wave (W2W) for marine robotics. The use of natural language allows real-time programming of subsea AUVs. We formulate novel language rules and intuitive command structures for efficient language-to-mission mapping. We develop a sequence-to-sequence learning pipeline based on {\tt T5-Small} model, which demonstrates robust and more efficient performance compared to other SLM/LLM architectures. We also developed a mission visualizer for users to validate the W2W-generated mission map before deployment. Through comprehensive quantitative, qualitative, and a user study to validate the W2W through real subsea deployment scenarios. In future W2W versions, we will incorporate human-machine dialogue and spontaneous question-answering for embodied mission reasoning. We also intend to explore language-driven HMIs and adapt W2W capabilities for subsea telerobotics applications.

\section*{Acknowledgements}
\vspace{-1mm}
This work is supported in part by the NSF grants \#$2330416$ and \#$2326159$. We are thankful to Dr. Arthur Trembanis, Dr. Herbert Tanner, and Dr. Kleio Baxevani at the University of Delaware for facilitating our field trials at the 2024 Autonomous Systems Bootcamp. We also acknowledge the SUS study participants for helping us conduct the UI evaluation.  

\small{
\bibliography{ref}

\begin{thebibliography}{10}

\bibitem{wang2024large}
J.~Wang, Z.~Wu, Y.~Li, H.~Jiang, P.~Shu, E.~Shi, H.~Hu, C.~Ma, Y.~Liu, X.~Wang, Y.~Yao, X.~Liu, H.~Zhao, Z.~Liu, H.~Dai, L.~Zhao, B.~Ge, X.~Li, T.~Liu, and S.~Zhang, ``{Large Language Models for Robotics: Opportunities, Challenges, and Perspectives},'' {\em arXiv preprint arXiv:2401.04334}, 2024.

\bibitem{zhang2023large}
C.~Zhang, J.~Chen, J.~Li, Y.~Peng, and Z.~Mao, ``{Large Language Models for Human-Robot Interaction: A Review},'' {\em Biomimetic Intelligence and Robotics}, p.~100131, 2023.

\bibitem{islam2018dynamic}
M.~J. Islam, M.~Ho, and J.~Sattar, ``{Dynamic Reconfiguration of Mission Parameters in Underwater Human-Robot Collaboration},'' in {\em IEEE International Conference on Robotics and Automation ({ICRA})}, pp.~1--8, 2018.

\bibitem{lucas2020survey}
N.~Lucas~Martinez, J.-F. Mart{\'\i}nez-Ortega, P.~Castillejo, and V.~Beltran~Martinez, ``{Survey of Mission Planning and Management Architectures for Underwater Cooperative Robotics Operations},'' {\em Applied Sciences}, vol.~10, no.~3, p.~1086, 2020.

\bibitem{sattar2009underwater}
J.~Sattar and G.~Dudek, ``{Underwater Human-robot Interaction via Biological Motion Identification},'' in {\em {Robotics: Science and Systems (RSS)}}, 2009.

\bibitem{islam2018understanding}
M.~J. Islam, M.~Ho, and J.~Sattar, ``{Understanding Human Motion and Gestures for Underwater Human–Robot Collaboration},'' {\em {Journal of Field Robotics (JFR)}}, vol.~36, no.~5, pp.~851--873, 2019.

\bibitem{antonio2017Nav}
A.~Vasilijevi{\'c}, D.~Nad, F.~Mandi{\'c}, N.~Mi{\v{s}}kovi{\'c}, and Z.~Vuki{\'c}, ``{Coordinated Navigation of Surface and Underwater Marine Robotic Vehicles for Ocean Sampling and Environmental Monitoring},'' {\em IEEE/ASME transactions on mechatronics}, vol.~22, no.~3, pp.~1174--1184, 2017.

\bibitem{gussen2021optimization}
C.~M. Gussen, C.~Laot, F.-X. Socheleau, B.~Zerr, T.~Le~M{\'e}zo, R.~Bourdon, and C.~Le~Berre, ``{Optimization of Acoustic Communication Links for a Swarm of AUVs: The COMET and NEMOSENS Examples},'' {\em Applied Sciences}, vol.~11, no.~17, p.~8200, 2021.

\bibitem{zhou2023review}
J.~Zhou, Y.~Si, and Y.~Chen, ``{A Review of Subsea AUV Technology},'' {\em Journal of Marine Science and Engineering}, vol.~11, p.~1119, 2023.

\bibitem{Kong2024_EmbodiedAI}
X.~Kong, W.~Zhang, J.~Hong, and T.~Braunl, ``Embodied ai in mobile robots: Coverage path planning with large language models,'' 2024.

\bibitem{lin2023development}
J.~Lin, H.~Gao, R.~Xu, C.~Wang, L.~Guo, and S.~Xu, ``{The Development of LLMs for Embodied Navigation},'' {\em arXiv preprint arXiv:2311.00530}, 2023.

\bibitem{park2023visual}
S.-M. Park and Y.-G. Kim, ``{Visual Language Navigation: A Survey and Open Challenges},'' {\em Artificial Intelligence Review}, vol.~56, no.~1, pp.~365--427, 2023.

\bibitem{shao2023prompting}
Z.~Shao, Z.~Yu, M.~Wang, and J.~Yu, ``{Prompting Large Language Models with Answer Heuristics for Knowledge-Based Visual Question Answering},'' in {\em Proceedings of the IEEE/CVF Conference on Computer Vision and Pattern Recognition}, pp.~14974--14983, 2023.

\bibitem{brooks2023instructpix2pix}
T.~Brooks, A.~Holynski, and A.~A. Efros, ``{Instructpix2pix: Learning to Follow Image Editing Instructions},'' in {\em Proceedings of the IEEE/CVF Conference on Computer Vision and Pattern Recognition}, pp.~18392--18402, 2023.

\bibitem{huang2022inner}
W.~Huang, F.~Xia, T.~Xiao, H.~Chan, J.~Liang, P.~Florence, A.~Zeng, J.~Tompson, I.~Mordatch, Y.~Chebotar, P.~Sermanet, N.~Brown, T.~Jackson, L.~Luu, S.~Levine, K.~Hausman, and B.~Ichter, ``{Inner Monologue: Embodied Reasoning Through Planning with Language Models},'' {\em arXiv preprint arXiv:2207.05608}, 2022.

\bibitem{yang2023oceanchat}
R.~Yang, M.~Hou, J.~Wang, and F.~Zhang, ``{OceanChat: Piloting Autonomous Underwater Vehicles in Natural Language},'' {\em arXiv preprint arXiv:2309.16052}, 2023.

\bibitem{bi2023oceangpt}
Z.~Bi, N.~Zhang, Y.~Xue, Y.~Ou, D.~Ji, G.~Zheng, and H.~Chen, ``{OceanGPT: A Large Language Model for Ocean Science Tasks},'' {\em arXiv preprint arXiv:2310.02031}, 2023.

\bibitem{Raffel2019ExploringTL}
C.~Raffel, N.~Shazeer, A.~Roberts, K.~Lee, S.~Narang, M.~Matena, Y.~Zhou, W.~Li, and P.~J. Liu, ``{Exploring the Limits of Transfer Learning with a Unified Text-to-Text Transformer},'' {\em Journal of Machine Learning Research}, vol.~21, no.~140, pp.~1--67, 2020.

\bibitem{andy2022subsea}
G.~S. Andy, L.~Mathieu, G.~Bill, and L.~Damian, ``{Subsea Robotics-AUV Pipeline Inspection Development Project},'' in {\em Offshore Technology Conference}, p.~D021S024R001, OTC, 2022.

\bibitem{NemoSens}
R.~Inc., ``{NemoSens: An Open-Architecture, Cost-Effective, and Modular Micro-AUV}.'' \url{https://rtsys.eu/nemosens-micro-auv}, 2020.

\bibitem{mcmahon2016mission}
J.~McMahon and E.~Plaku, ``{Mission and Motion Planning for Autonomous Underwater Vehicles Operating in Spatially and Temporally Complex Environments},'' {\em IEEE Journal of Oceanic Engineering}, vol.~41, no.~4, pp.~893--912, 2016.

\bibitem{achiam2023gpt}
J.~Achiam, S.~Adler, S.~Agarwal, L.~Ahmad, I.~Akkaya, F.~L. Aleman, D.~Almeida, J.~Altenschmidt, S.~Altman, S.~Anadkat, {\em et~al.}, ``{Gpt-4 Technical Report},'' {\em arXiv preprint arXiv:2303.08774}, 2023.

\bibitem{palnitkar2023chatsim}
A.~Palnitkar, R.~Kapu, X.~Lin, C.~Liu, N.~Karapetyan, and Y.~Aloimonos, ``{ChatSim: Underwater Simulation with Natural Language Prompting},'' in {\em OCEANS 2023-MTS/IEEE US Gulf Coast}, pp.~1--7, IEEE, 2023.

\bibitem{bartlarge}
M.~Lewis, ``{BART: Denoising Sequence-to-Sequence Pre-training for Natural Language Generation, Translation, and Comprehension},'' {\em arXiv preprint arXiv:1910.13461}, 2019.

\bibitem{mariannmt}
M.~Junczys-Dowmunt, R.~Grundkiewicz, T.~Dwojak, H.~Hoang, K.~Heafield, T.~Neckermann, F.~Seide, U.~Germann, A.~Fikri~Aji, N.~Bogoychev, A.~F.~T. Martins, and A.~Birch, ``{Marian: Fast Neural Machine Translation in C++},'' in {\em Proceedings of ACL, System Demonstrations}, pp.~116--121, 2018.

\bibitem{brooke1996sus}
J.~Brooke, ``{SUS- A Quick and Dirty Usability Scale},'' {\em Usability Evaluation in Industry}, vol.~189, no.~194, pp.~4--7, 1996.

\bibitem{chandarana2017fly}
M.~Chandarana, E.~L. Meszaros, A.~Trujillo, and B.~D. Allen, ``{'Fly Like This': Natural Language Interface for UAV Mission Planning},'' in {\em International Conference on Advances in Computer-Human Interactions}, no.~NF1676L-26108, 2017.

\bibitem{garcia2019high}
S.~Garc{\'\i}a, P.~Pelliccione, C.~Menghi, T.~Berger, and T.~Bures, ``{High-Level Mission Specification for Multiple Robots},'' in {\em Proceedings of the 12th ACM SIGPLAN International Conference on Software Language Engineering}, pp.~127--140, 2019.

\bibitem{shah2021ving}
D.~Shah, B.~Eysenbach, G.~Kahn, N.~Rhinehart, and S.~Levine, ``{Ving: Learning Open-World Navigation with Visual Goals},'' in {\em IEEE International Conference on Robotics and Automation (ICRA)}, pp.~13215--13222, 2021.

\bibitem{huang2019visual}
G.~Huang, ``{Visual-Inertial Navigation: A Concise Review},'' in {\em International Conference on Robotics and Automation (ICRA)}, pp.~9572--9582, IEEE, 2019.

\bibitem{chandarana2018challenges}
M.~Chandarana, E.~L. Meszaros, A.~Trujillo, and B.~D. Allen, ``{Challenges of Using Gestures in Multimodal HMI for Unmanned Mission Planning},'' in {\em Proceedings of the AHFE 2017 International Conference on Human Factors in Robots and Unmanned Systems}, Springer, 2018.

\bibitem{shah2011language}
M.~A. Shah, ``{A Language-Based Software Framework for Mission Planning in Autonomous Mobile Robots},'' Master's thesis, Pennsylvania State University, 2011.

\bibitem{mutschler2005language}
D.~W. Mutschler, ``{Language Based Simulation, Flexibility, and Development Speed in the Joint Integrated Mission Model},'' in {\em Proceedings of the Winter Simulation Conference.}, p.~8, IEEE, 2005.

\bibitem{silva2014development}
D.~C. Silva, P.~H. Abreu, L.~P. Reis, and E.~Oliveira, ``{Development of a Flexible Language for Mission Description for Multi-Robot Missions},'' {\em Information Sciences}, vol.~288, pp.~27--44, 2014.

\bibitem{trujillo2017using}
A.~C. Trujillo, J.~Puig-Navarro, S.~B. Mehdi, and A.~K. McQuarry, ``{Using Natural Language to Enable Mission Managers to Control Multiple Heterogeneous UAVs},'' in {\em Proceedings of the AHFE 2016 International Conference on Human Factors in Robots and Unmanned Systems}, pp.~267--280, Springer, 2017.

\bibitem{cividanes2021extended}
F.~D.~S. Cividanes, M.~G.~V. Ferreira, and F.~de~Novaes~Kucinskis, ``{An Extended HTN Language for Onboard Planning and Acting Applied to a Goal-Based Autonomous Satellite},'' {\em IEEE Aerospace and Electronic Systems Magazine}, vol.~36, no.~8, pp.~32--50, 2021.

\bibitem{Cui2024_UAVTaskPlanning}
J.~Cui, G.~Liu, H.~Wang, Y.~Yu, and J.~Yang, ``Tpml: Task planning for multi-uav system with large language models,'' in {\em 2024 IEEE 18th International Conference on Control \& Automation (ICCA)}, pp.~886--891, 2024.

\bibitem{wu2023tidybot}
J.~Wu, R.~Antonova, A.~Kan, M.~Lepert, A.~Zeng, S.~Song, J.~Bohg, S.~Rusinkiewicz, and T.~Funkhouser, ``{Tidybot: Personalized Robot Assistance with Large Language Models},'' {\em Autonomous Robots}, vol.~47, no.~8, pp.~1087--1102, 2023.

\bibitem{Nazarov2025_SLM}
B.~Nazarov, D.~Frolova, Y.~Lubarsky, A.~Gaissinski, and P.~Kisilev, ``Rethinking data: Towards better performing domain-specific small language models,'' 2025.

\bibitem{kwon2024language}
T.~Kwon, N.~Di~Palo, and E.~Johns, ``{Language Models as Zero-Shot Trajectory Generators},'' {\em IEEE Robotics and Automation Letters}, 2024.

\bibitem{Thawakar2024MobiLlama}
O.~Thawakar, A.~Vayani, S.~Khan, H.~Cholakal, R.~M. Anwer, M.~Felsberg, T.~Baldwin, E.~P. Xing, and F.~S. Khan, ``Mobillama: Towards accurate and lightweight fully transparent gpt,'' 2024.

\bibitem{shah2023lm}
D.~Shah, B.~Osi{\'n}ski, B.~Ichter, and S.~Levine, ``{LM-Nav: Robotic Navigation with Large Pre-trained Models of Language, Vision, and Action},'' in {\em Conference on Robot Learning}, pp.~492--504, 2023.

\bibitem{gao2024physically}
J.~Gao, B.~Sarkar, F.~Xia, T.~Xiao, J.~Wu, B.~Ichter, A.~Majumdar, and D.~Sadigh, ``{Physically Grounded Vision-Language Models for Robotic Manipulation},'' in {\em IEEE International Conference on Robotics and Automation (ICRA)}, pp.~12462--12469, 2024.

\bibitem{mandi2024roco}
Z.~Mandi, S.~Jain, and S.~Song, ``{Roco: Dialectic Multi-robot Collaboration with Large Language Models},'' in {\em IEEE International Conference on Robotics and Automation (ICRA)}, pp.~286--299, 2024.

\bibitem{sanchez2020autonomous}
P.~J.~B. S{\'a}nchez, M.~Papaelias, and F.~P.~G. M{\'a}rquez, ``{Autonomous Underwater Vehicles: Instrumentation and Measurements},'' {\em IEEE Instrumentation \& Measurement Magazine}, vol.~23, pp.~105--114, 2020.

\bibitem{wei2021architecture}
D.~Wei, P.~Hao, and H.~Ma, ``{Architecture Design and Implementation of UUV Mission Planning},'' in {\em 2021 40th Chinese Control Conference (CCC)}, pp.~1869--1873, IEEE, 2021.

\bibitem{mcmanus2005multidisciplinary}
I.~A. McManus, {\em {A Multidisciplinary Approach to Highly Autonomous UAV Mission Planning and Piloting for Civilian Airspace}}.
\newblock PhD thesis, Queensland University of Technology, 2005.

\bibitem{oliveira1998mission}
P.~Oliveira, A.~Pascoal, V.~Silva, and C.~Silvestre, ``{Mission Control of the MARIUS Autonomous Underwater Vehicle: System Design, Implementation, and Sea trials},'' {\em International journal of systems science}, vol.~29, no.~10, pp.~1065--1080, 1998.

\bibitem{gonzalez2020autonomous}
J.~Gonz{\'a}lez-Garc{\'\i}a, A.~G{\'o}mez-Espinosa, E.~Cuan-Urquizo, L.~G. Garc{\'\i}a-Valdovinos, T.~Salgado-Jim{\'e}nez, and J.~A. Escobedo~Cabello, ``{Autonomous Underwater Vehicles: Localization, Navigation, and Communication for Collaborative Missions},'' {\em Applied Sciences}, vol.~10, no.~4, p.~1256, 2020.

\bibitem{whitt2020future}
C.~Whitt, J.~Pearlman, B.~Polagye, F.~Caimi, F.~Muller-Karger, A.~Copping, H.~Spence, S.~Madhusudhana, W.~Kirkwood, L.~Grosjean, {\em et~al.}, ``{Future Vision for Autonomous Ocean Observations},'' {\em Frontiers in Marine Science}, vol.~7, p.~697, 2020.

\bibitem{enan2022robotic}
S.~S. Enan, M.~Fulton, and J.~Sattar, ``{Robotic Detection of a Human-comprehensible Gestural Language for Underwater Multi-human-robot Collaboration},'' in {\em 2022 IEEE/RSJ International Conference on Intelligent Robots and Systems (IROS)}, pp.~3085--3092, IEEE, 2022.

\bibitem{abdullah2024ego2exo}
A.~Abdullah, R.~Chen, I.~Rekleitis, and M.~J. Islam, ``{Ego-to-Exo: Interfacing Third Person Visuals from Egocentric Views in Real-time for Improved ROV Teleoperation},'' in {\em In Review. ArXiv: 2407.00848}, 2024.

\bibitem{xu2021vr}
F.~Xu, Q.~Zhu, S.~Li, Z.~Song, and J.~Du, ``{VR-Based Haptic Simulator for Subsea Robot Teleoperations},'' in {\em Computing in Civil Engineering}, pp.~1024--1032, 2021.

\bibitem{hallin2009using}
N.~J. Hallin, B.~Johnson, H.~Egbo, M.~O'Rourke, and D.~Edwards, ``{Using Language-Centered Intelligence to Optimize Mine-Like Object Inspections for a Fleet of Autonomous Underwater Vehicles},'' {\em UUST'09}, 2009.

\bibitem{Dudek2007_RoboChat}
G.~Dudek, J.~Sattar, and A.~Xu, ``A visual language for robot control and programming: A human-interface study,'' in {\em Proceedings 2007 IEEE International Conference on Robotics and Automation}, pp.~2507--2513, 2007.

\bibitem{Xu2008_RoboChatGest}
A.~Xu, G.~Dudek, and J.~Sattar, ``A natural gesture interface for operating robotic systems,'' in {\em 2008 IEEE International Conference on Robotics and Automation}, pp.~3557--3563, 2008.

\bibitem{Capocci2017_ROVReview}
R.~Capocci, G.~Dooly, E.~Omerdi{\'c}, J.~Coleman, T.~Newe, and D.~Toal, ``{Inspection-Class Remotely Operated Vehicles—A Review},'' {\em Journal of Marine Science and Engineering}, vol.~5, no.~1, p.~13, 2017.

\bibitem{VideoRay}
VideoRay, ``{OctoView: A Cutting Edge Mixed Reality Interface for Pilots and Teams}.'' \url{https://videoray.com/labs/}, 2024.

\bibitem{xu2024augmented}
F.~Xu, T.~Nguyen, and J.~Du, ``{Augmented Reality for Maintenance Tasks with ChatGPT for Automated Text-to-Action},'' {\em Journal of Construction Engineering and Management}, vol.~150, no.~4, 2024.

\bibitem{laranjeira20203d}
M.~Laranjeira, A.~Arnaubec, L.~Brignone, C.~Dune, and J.~Opderbecke, ``{3D Perception and Augmented Reality Developments in Underwater Robotics for Ocean Sciences},'' {\em Current Robotics Reports}, vol.~1, pp.~123--130, 2020.

\bibitem{islam2024eob}
M.~J. Islam, ``{Eye On the Back: Augmented Visuals for Improved ROV Teleoperation in Deep Water Surveillance and Inspection},'' in {\em SPIE Defense and Commercial Sensing}, (Maryland, USA), SPIE, 2024.

\bibitem{gao2020mission}
F.~Gao, ``{Mission Planning and Replanning for ROVs},'' Master's thesis, NTNU, 2020.

\bibitem{manley2018aquanaut}
J.~E. Manley, S.~Halpin, N.~Radford, and M.~Ondler, ``{Aquanaut: A New Tool for Subsea Inspection and Intervention},'' in {\em OCEANS 2018 MTS/IEEE Charleston}, pp.~1--4, IEEE, 2018.

\bibitem{parekh2023underwater}
P.~Parekh, C.~McGuire, and J.~Imyak, ``Underwater robotics semantic parser assistant,'' {\em arXiv preprint arXiv:2301.12134}, 2023.

\bibitem{samuelson2024guided}
C.~R. Samuelson and J.~G. Mangelson, ``{A Guided Gaussian-Dirichlet Random Field for Scientist-in-the-Loop Inference in Underwater Robotics},'' in {\em 2024 IEEE International Conference on Robotics and Automation (ICRA)}, pp.~9448--9454, IEEE, 2024.

\bibitem{rankin2021robotic}
I.~C. Rankin, S.~McCammon, and G.~A. Hollinger, ``{Robotic Information Gathering Using Semantic Language Instructions},'' in {\em 2021 IEEE International Conference on Robotics and Automation (ICRA)}, pp.~4882--4888, IEEE, 2021.

\bibitem{potokar2022holoocean}
E.~Potokar, S.~Ashford, M.~Kaess, and J.~G. Mangelson, ``{HoloOcean: An Underwater Robotics Simulator},'' in {\em International Conference on Robotics and Automation (ICRA)}, pp.~3040--3046, IEEE, 2022.

\bibitem{manhaes2016uuv}
M.~M.~M. Manh{\~a}es, S.~A. Scherer, M.~Voss, L.~R. Douat, and T.~Rauschenbach, ``{UUV Simulator: A Gazebo-Based Package for Underwater Intervention and Multi-Robot Simulation},'' in {\em Oceans MTS/IEEE Monterey}, pp.~1--8, 2016.

\bibitem{prats2012open}
M.~Prats, J.~Perez, J.~J. Fern{\'a}ndez, and P.~J. Sanz, ``{An Open Source Tool for Simulation and Supervision of Underwater Intervention Missions},'' in {\em IEEE/RSJ International Conference on Intelligent Robots and Systems}, pp.~2577--2582, 2012.

\bibitem{Leaflet}
V.~Agafonkin, ``{Leaflet}.'' \url{https://leafletjs.com/}, 2010.

\bibitem{devlin2018bert}
J.~Devlin, ``{BERT: Pre-training of Deep Bidirectional Transformers for Language Understanding},'' {\em arXiv preprint arXiv:1810.04805}, 2018.

\bibitem{TransformerBasedNewsSummarization}
A.~Choudhary, M.~Alugubelly, and R.~Bhargava, ``{A Comparative Study on Transformer-based News Summarization},'' in {\em International Conference on Developments in eSystems Engineering (DeSE)}, pp.~256--261, 2023.

\bibitem{BLEU}
K.~Papineni, S.~Roukos, T.~Ward, and W.-J. Zhu, ``{BLEU: A Method for Automatic Evaluation of Machine Translation},'' in {\em Proceedings of the 40th Annual Meeting of the Association for Computational Linguistics}, pp.~311--318, 2002.

\bibitem{METEOR}
S.~Banerjee and A.~Lavie, ``{METEOR: An Automatic Metric for MT Evaluation with Improved Correlation with Human Judgments},'' in {\em Proceedings of the ACL Workshop on Intrinsic and Extrinsic Evaluation Measures for Machine Translation and/or Summarization}, pp.~65--72, 2005.

\bibitem{Huang2025_LLMHallucinations}
L.~Huang, W.~Yu, W.~Ma, W.~Zhong, Z.~Feng, H.~Wang, Q.~Chen, W.~Peng, X.~Feng, B.~Qin, and T.~Liu, ``A survey on hallucination in large language models: Principles, taxonomy, challenges, and open questions,'' {\em ACM Transactions on Information Systems}, vol.~43, p.~1–55, Jan. 2025.

\bibitem{2024Vosk}
M.~H. Thekiya, ``vosk-api.'' \url{https://github.com/alphacep/vosk-api}, 2024.

\end{thebibliography}
}

\end{document}